\documentclass[10pt,twocolumn,letterpaper]{article}

\usepackage{iccv}
\usepackage{times}
\usepackage{epsfig}
\usepackage{graphicx}
\usepackage{amsmath}
\usepackage{amssymb}
\usepackage{algorithm}
\usepackage{algpseudocode}

\usepackage[breaklinks=true,bookmarks=false]{hyperref}

\iccvfinalcopy 

\def\iccvPaperID{24} 

\ificcvfinal\fi

\begin{document}

\title{Logarithm-transform aided Gaussian Sampling for Few-Shot Learning}

\author{Vaibhav Ganatra\\
BITS Pilani, Goa\\
Goa, India\\
{\tt\small f20190010@goa.bits-pilani.ac.in}
}

\maketitle
\ificcvfinal\thispagestyle{empty}\fi

\begin{abstract}
Few-shot image classification has recently witnessed the rise of representation learning being utilised for models to adapt to new classes using only a few training examples. Therefore, the properties of the representations, such as their underlying probability distributions, assume vital importance. Representations sampled from Gaussian distributions have been used in recent works, \cite{free-lunch} to train classifiers for few-shot classification. These methods rely on transforming the distributions of experimental data to approximate Gaussian distributions for their functioning. In this paper, I propose a novel Gaussian transform, that outperforms existing methods on transforming experimental data into Gaussian-like distributions. I then utilise this novel transformation for few-shot image classification and show significant gains in performance, while sampling lesser data.
\end{abstract}

\section{Introduction}
Learning from limited data, and adapting neural networks to unforeseen tasks has attracted signification attention in recent years. This is essential since the development of large datasets for supervised training requires significant costs, in terms of finances and the amount of human effort required. Considerable progress has been made in machine learning in the limited data regime. After the development of sophisticated optimization-based meta-learning techniques such as Model-agnostic-Meta-Learning \cite{maml}, and metric-based meta-learning techniques, such as ProtoNets and MatchingNets\cite{protonet,nk-shot}, there has been a recent shift towards representation learning for few-shot learning \cite{good-representation, metafree, compositional-representations, global-representations}. For example, Tian \textit{et al.} \cite{good-representation} leverage self-supervision and regularization for learning meaningful representations, and state that ``a good embedding is all you need" for few-shot image classification. They propose a meta-free model as a new State-of-the-Art for few-shot image classification. Therefore, studying the properties (such as the underlying probability distributions), of the learned representations is a useful pursuit in decoding their usefulness in few-shot learning.



Gaussian representations play an essential role in few-shot learning. For example, instead of using point prototypes for few-shot image classification, Lin \textit{et al.} \cite{Gaussian-prototypes} propose modelling prototypes as multi-dimensional Gaussian distributions, and rectify these prototypes using mutual information maximization. Yang \textit{et al.} \cite{free-lunch} propose a distribution calibration mechanism to calibrate the representations of few-shot data and train classifiers by sampling Gaussian data around the calibrated representations. They make use of Tukey's Ladder of Powers \cite{tukey} to transform the data so that the underlying distribution of the transformed data is approximately Gaussian. 

In this paper, I propose a novel ``data-to-Gaussian" transform that ``Gaussianizes" data. (Gaussianization of data or inducing ``normality" into data refers to transforming the data so that its underlying distribution is approximately normal).
I demonstrate the utility of the proposed transformation method by transforming a variety of data distributions (including data from experimental datasets) and show (through qualitative and quantitative evaluation) that the transform produces data following a distribution that is closer to a Gaussian distribution. Finally, I replace the distribution calibration mechanism proposed by Yang \textit{et al.} \cite{free-lunch} with the proposed transformation method and show small yet consistent improvements in the classification accuracies, while sampling a significantly lesser amount of data.

Therefore, in this paper, I make the following contributions - 
\begin{itemize}
    \item I propose a novel transformation (called the Log-Tukey transformation) to induce ``Gaussianization" within experimental data.
    \item I replace the distribution calibration mechanism proposed by Yang \textit{et al.} \cite{free-lunch} with the novel transform and devise an alternative algorithm for few-shot learning.
    \item I perform experiments on commonly known benchmark datasets, and show significant performance gains while sampling 5x lesser datapoints.
\end{itemize}

The rest of the paper is organized as follows - Sec.\ref{related-work} details the related work, both in terms of making data more Gaussian, and representation learning for few-shot learning. Sec.\ref{log-tukey-transform-section} explains the Log-Tukey transform for Gaussianization of data. In Sec.\ref{Gaussian-sampling}, I first briefly discuss the distribution calibration method proposed by Yang \textit{et al.} \cite{free-lunch}, since the proposed work is heavily based on it. Next, I detail how I incorporate the Log-Tukey transform into the algorithmic setup proposed by them. Sec.\ref{experiments} shows experimental results of how the method proposed in Sec.\ref{Gaussian-sampling} outperforms the baseline and Sec.~\ref{conclusion} concludes the paper.
\section{Related Work}
\label{related-work}

\textbf{Making data Gaussian-like} - A significant amount of prior work exists in transforming experimental data such that it follows Gaussian-like distributions. This is because of the usefulness of Gaussian data. Tukey's Ladders of Powers \cite{tukey} are one of the prominent methods available for this task. Other transforms such as Log transforms and Inverse transforms are also used for this purpose. In addition to these, the Box-Cox \cite{box-cox} transform is another useful method that is widely used for transforming data and making it more Gaussian-like, but can only be used for positive values. The Yeo-Johnson transform \cite{yeo-johnson} is a modification of the Box-Cox transform and can also be used for negative values.

\textbf{Representation Learning for Few-Shot Learning} - Tian \textit{et al.} \cite{good-representation} debunk the need for complicated meta-learning methods for few-shot learning, and emphasize the utility of representations learned through a proxy task such as image classification. Mangla \textit{et al.} \cite{feature-extractor} make use of self-supervision and regularization to learn meaningful structures in the representations of data. Luo \textit{et al.} \cite{global-representations} approach few-shot learning by learning global representations whereas Tokmakov \textit{et al.} \cite{compositional-representations} learn compositional representations for few-shot learning.


\section{Log-Tukey Transform}
\label{log-tukey-transform-section}
As stated earlier, “Gaussianization” of data plays an essential role in few-shot learning. 
Hence, there exist multiple techniques to transform existing data such that it follows an approximate Gaussian distribution. Tukey’s Ladder of Powers transform \cite{tukey}, as shown in Eq.~\ref{tukey-eq}, is by far one of the most popular methods used to Gaussianize data.

\begin{equation}
\label{tukey-eq}
    \hat{x} = \begin{cases}
        x^{\lambda} , if \lambda \neq 0\\
        log(x), if \lambda = 0
    \end{cases}
\end{equation}
Commonly, the value of $\lambda$ used is 0.5. However, only using exponents of the data is not immune to data skew and does not ensure maximum ``normality"/``Gaussianization" in the data.

\begin{figure}[b]
    \centering
    \includegraphics[width = 6cm]{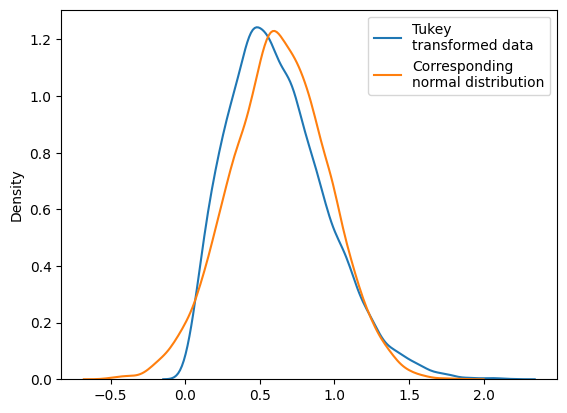}
    \caption{KDE plot (Probability distribution function) of Tukey-transformed data sampled from an Exponential(0.5) distribution. The ``Corresponding normal distribution" shown in the figure is a Gaussian distribution with the same mean and variance as the Tukey-transformed data. Due to skew in the data, the transformed data is somewhat different from the corresponding normal distribution.}
    \label{tukey-skew}
\end{figure}

Fig.~\ref{tukey-skew} shows the probability distribution function of a data sample drawn from an Exponential(0.5) distribution after transforming it using Tukey's Square Root transform and that of the corresponding Gaussian distribution.
The peaks of the Tukey-Transformed data distribution and the corresponding Gaussian distribution are misaligned (The peak of the Tukey-transformed data is to the left of the peak of the Gaussian distribution). This happens because the exponential distribution is positively skewed and the Tukey-transform is unable to sufficiently shrink the large values, consequently, the distribution of the transformed data is still positively skewed. (Gaussian distributions do not have any skew)
This is a common problem with the method, therefore, maximum ``normality" is not ensured in the transformed data. A transform that is used to Gaussianize data must significantly reduce/remove the skew in a skewed distribution. Skew (and long tails) in the data is often overcome by using the logarithm function \cite{log-transform}. Consequently, in this paper, I introduce the Log-Tukey transform, which makes use of logarithm along with Tukey's Square-root transform, as shown in Eq.~\ref{log-tukey-transform}.

\begin{equation}
\label{log-tukey-transform}
    \hat{x} = log ( \sqrt{x} + \epsilon + 1)
\end{equation}

 where $\epsilon$ is a small value to prevent the transformation from zeroing out the input. In the experiments, a value of 1e-4 is used for $\epsilon$. The $+1$ is added to ensure that the resulting values after the transform are positive.

\begin{figure}[t]
    \centering
    \includegraphics[width = 6cm]{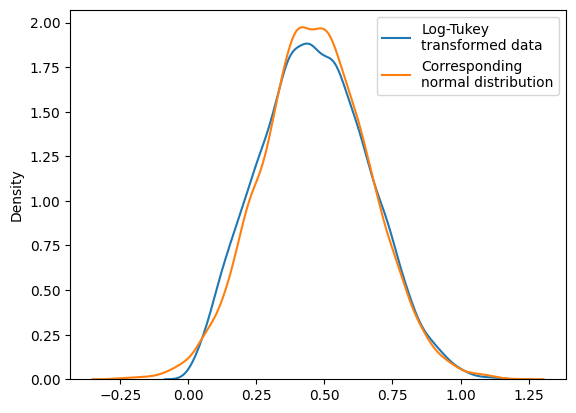}
    \caption{KDE plot (Probability distribution function) of Log-tukey-transformed data sampled from an Exponential(0.5) distribution. The ``Corresponding normal distribution" shown in the figure is a Gaussian distribution with the same mean and variance as the Log-Tukey transformed data. The skew in the data is removed and the distribution is much closer to a Gaussian distribution with the same mean and variance.}
    \label{log-tukey-no-bias}
\end{figure}
The Log-Tukey transform removes the data skew and shrinks the data in a way that the underlying distribution is much closer to a Gaussian-distribution, with the same mean and variance, as shown in Fig.~\ref{log-tukey-no-bias} (The peaks are horizontally aligned, and the edges lie very close). A quantitative comparison of the various ``Gaussianization" methods is carried out, where the “closeness” of the distributions is estimated in terms of the Wasserstein distance \cite{wasserstein}, which can be interpreted as the minimum cost of transforming one of the distributions into the other. Here, I also consider the Box-Cox \cite{box-cox} and Yeo-Johnson \cite{yeo-johnson} transforms, which are commonly used for Gaussianization of data. As is evident from Table.~\ref{w-distance}, the Log-Tukey transformation has the lowest Wasserstein distance in all cases, thereby showing the enhanced ability of the transformation in inducing ``normality" within the data ( The table also includes data from the Iris dataset \cite{iris}). In deep learning, this ability is crucial in places where we do not know the ground truth distributions of data/weights, but assume that the underlying distribution is Gaussian. An example of this is shown in Sec.\ref{Gaussian-sampling} where we sample data for few-shot learning in the representation space, assuming it follows a multivariate Gaussian distribution.

\begin{table}[]
    \centering
    \begin{tabular}{|c|c|c|c|c|}
    \hline
    Distribution & Transform & Mean & Std & Wasserstein\\
    &&&Dev&Distance \textcolor{green}{$\downarrow$} \\
    \hline
    Uni(0,1) & None & 0.5 & 0.289 & 0.0458\\
    Uni(0,1) & Tukey & 0.661 & 0.238 & 0.0402\\
    Uni(0,1) & Box-Cox & -0.598 & 0.384 & 0.0579\\
    Uni(0,1) & Yeo-Johnson & 0.456 & 0.258 & 0.0405\\
    Uni(0,1) & \textbf{Log-tukey} & 0.497 & 0.152 & \textbf{0.0308}\\
    \hline
    Exp(0.5) & None & 0.5 & 0.5 & 0.1482\\
    Exp(0.5) & Tukey & 0.625 & 0.326 & 0.0338\\
    Exp(0.5) & Box-Cox & -0.928 & 0.826 & 0.0227\\
    Exp(0.5) & Yeo-Johnson & 0.244 & 0.145 & 0.0169\\
    Exp(0.5) & \textbf{Log-Tukey} & 0.466 & 0.197 & \textbf{0.008}\\
    \hline
    Feature 0 & None & 5.843 & 0.825 & 0.0827\\
    from & Tukey & 2.411 & 0.17 & 0.0153\\
    Iris & Box-Cox& 1.549 & 0.109 & 0.01\\
    Dataset & Yeo-Johnson & 1.43 & 0.065 & 0.0057\\
     & \textbf{Log-Tukey} & 1.226 & 0.05 & \textbf{0.0042}\\
     \hline

    \end{tabular}
    \caption{Quantitative comparison of "Gaussianization" methods. Uni stands for the continuous uniform distribution, and Exp stands for the exponential distribution.}
    \label{w-distance}
\end{table}

\section{Gaussian Sampling for Few-Shot Classification}
\label{Gaussian-sampling}

In this section, I make use of the Log-Tukey transform for few-shot image classification. 

\subsection{Problem Setting}
The typical few-shot classification problem formulation is adopted, where we have a labelled dataset $D = \{(x^i,y^i) | 1 \leq i \leq T \}$, where $x^i$ is a data sample, and $y^i$ is the corresponding label, $T$ denotes the size of the dataset. Each datapoint in the dataset is labelled as one of $|C|$ classes, where $C$ denotes the set of all classes. $C$ is partitioned into base classes $C_b$ and novel classes $C_n$, such that $C_b \cap C_n = \phi$ and $C_b \cup C_n = C$. The few-shot classification model is trained on the base classes, and the goal is to train the model in such a manner that it is able to adapt well to the novel classes, using only a few examples. Typically, abundant data is available for the base classes but only few samples are available for the novel classes. The generalization ability of the model is evaluated in terms of accuracy in N-way-K-shot tasks \cite{nk-shot}, where each task consists of $N$ novel classes sampled from $C_n$ and $K$ labeled samples from each of the $N$ classes. This few-shot labeled set available for model adaption is called the support set. The performance of the model is evaluated on the query set $Q$, which consists of $q$ test cases from each of the $N$ classes $Q = {(x^i, y^i)}^{N \cdot K+N ×q}_{i=N \cdot K+1}$. Therefore, the average accuracy of the model on multiple N-way-K-shot tasks is used as an estimate of the model performance.


\subsection{Distribution Calibration \cite{free-lunch}}
In accordance with the growing interest in using effective representations for few-shot learning , Yang \textit{et al.} \cite{free-lunch} propose a distribution calibration mechanism which uses statistics of the well-separated base-classes to calibrate the representations of novel classes, and make them separable in a few-shot setting, without over-fitting.
They utilize the training method proposed by Mangla \textit{et al.} \cite{feature-extractor} trained on the base classes, as a feature extractor $F$. After training, they record the classwise statistics - mean $\mu_{base} = \{\mu_i | 1 \leq i \leq |C_b|\}$ and covariances {$\Sigma_{base} = \{\Sigma_i| 1 \leq i \leq |C_b|\}$ for the base-classes.

For each few-shot task, they calibrate the Tukey-transformed representation (obtained by using $F$ on the input image, followed by transformation using Eq.~\ref{tukey-eq}) of each image, using the base class statistics $\mu_{base}$ and $\Sigma_{base}$. Finally, they sample data around the calibrated representations, and train a linear classifier on data sampled around all points in the support set. They show that logistic regression/SVM classifiers trained on the calibrated representations and sampled data outperform sophisticated optimization, metric and generation-based meta-learning methods. One point that must be noted is that they sample data around each point in the support set, therefore, if $p$ points are sampled for each image in an N-way-K-shot task, a total of $N$ x $K$ x $p$ points are sampled. Algorithm.\ref{dc-algorithm} outlines the distribution calibration mechanism proposed by Yang \textit{et al.} \cite{free-lunch}.

\begin{algorithm}
    \caption{Algorithm for training a Few-Shot classifier using Distribution Calibration}   
    \textbf{Require} : Support features $S = (x,y)_{i=1}^{NxK}$\\
    \textbf{Require} : Base class statistics $\mu_{base}, \Sigma_{base}$

    1: Transform $S$ with the Tukey transform. (Eq.~\ref{tukey-eq})\\
    2: \textbf{for} $(x_i, y_i) \in S$ \textbf{do}\\
    3: ~~ Obtain calibrated mean $x_i'$ and covariance $\Sigma_i'$ for $x_i$ using the method proposed by Yang \textit{et al.} \cite{free-lunch}\\
    4: ~~ Sample multivariate Gaussian data using $x_i'$ and $\Sigma_i'$, and label them as $y_i$\\
    5: \textbf{end for}\\
    6: Train a linear classifier on the sampled + support set features
\label{dc-algorithm}
\end{algorithm}

\subsection{Gaussian Sampling}
I build on the work of Yang \textit{et al.} \cite{free-lunch}, since they utilize Tukey's Square Root transform. The experimental setting is a 5-way-5-shot image classifier. 
As is shown in Sec.~\ref{log-tukey-transform-section}, the Tukey-transform alone is not optimal in inducing ``normality" into the data. The sampled Gaussian data, therefore, is not as close to the ground truth representations of the novel classes, as possible and there is room for further correction.

Hence, I replace the Tukey-transform and distribution calibration steps with the Log-Tukey transformation on the representations of the novel classes in an attempt to make them more ``Gaussian", following which classwise means and covariances are calculated using the images from the support set. Next, data is sampled for each class around the mean and a linear classifier is  trained on the sampled data. Therefore, if $p$ points are sampled around each mean, a total of $N$ x $p$ datapoints are sampled, $p$ datapoints for each of the $N$ classes. In an N-way-K-shot setting, this is $K$ times lesser than that of the distribution calibration proposed by Yang \textit{et al.} \cite{free-lunch}. Since a 5-shot setting has been adopted, 5x lesser data is sampled. 

Thus, applying the Log-Tukey transform ensures that the data is more ``Gaussian", therefore sampling Gaussian data around the class-mean generates more accurate representations of the novel class. This ``Gaussian Sampling" of data from each class aids in few-shot learning and delivers a small yet consistent improvement in performance (as shown in Sec.~\ref{results}), while sampling 5x fewer data, thereby decreasing the computation. Algorithm.~\ref{GS-algorithm} shows the overall algorithm of training a few-shot classifier using Gaussian Sampling.


\begin{algorithm}
    \caption{Algorithm for training a Few-Shot classifier using Gaussian Sampling}   
    \textbf{Require} : Support features $S = (x,y)_{i=1}^{NxK}$

    1: Transform $S$ with the Log-Tukey transform (Eq.~\ref{log-tukey-transform})\\
    2: Calculate classwise-means $\mu$'s and covariance $\Sigma$'s from the transformed features\\
    3: Sample multivariate Gaussian data using $\mu$'s and $\Sigma$'s for each class, label them with the corresponding class label\\
    4: Train a linear classifier on the sampled + support set features
\label{GS-algorithm}
\end{algorithm}

\section{Experiments}
\label{experiments}

\subsection{Datasets}
I validate the Gaussian Sampling method on the miniImageNet \cite{miniImagenet} and the CUB \cite{cub} datasets. A variety of classes including various animals and objects can be found in the miniImageNet dataset while CUB is a more fine-grained dataset that includes various species of birds. Datasets with different levels of granularity may have different distributions for their feature space. I show the validity of the sampling mechanism on both datasets

\textbf{miniImageNet} is derived from ILSVRC-12 dataset \cite{imagenet}. It has 100 diverse classes with 600 images per class, each of size 84 × 84 × 3. The data split used in the following experiments is as proposed by Ravi \textit{et al.} \cite{miniImagenet}, with 64 base classes, 16 validation classes, and 20 novel classes.

\textbf{CUB} is a fine-grained few-shot classification benchmark. It has a total of 11,788 images, each of size 84 × 84 × 3, of 200 different classes of birds. The dataset is split into 100 base classes, 50 validation classes, and 50 novel classes, following Chen \textit{et al.}\cite{cub-split}

\subsection{Implementation and Metrics}
I follow the implementation provided by Yang \textit{et al.} \cite{free-lunch}. I use the method proposed by Mangla \textit{et al.} \cite{feature-extractor} as the feature-extractor, trained on the base classes and evaluate its performance on the novel classes. I adopt the 5-way-5-shot setting for the experiments. The values of hyper-parameters are as per the implementation provided by Yang \textit{et al.} \cite{free-lunch}. I evaluate the performance in terms of classification accuracy on the query set. This is evaluated over a total of 500 tasks, in 5 runs of sets of 100 tasks each. Since the distribution calibration proposed by Yang \textit{et al.} \cite{free-lunch} already surpasses the performance of other existing methods, and since the proposed method is heavily based on it, I only compare the proposed method against theirs \cite{free-lunch} \footnote{The code for the experiments is publicly available at - https://github.com/ganatra-v/gaussian-sampling-fsl}. I show the improvement in the performance across multiple runs for both the miniImageNet and CUB datasets.

\subsection{Results}
\label{results}

Table.~\ref{miniImageNet-runs} shows the accuracies of the distribution calibration mechanism \cite{free-lunch} and the Gaussian sampling (ours) mechanism for the miniImageNet dataset. In all the trials, the average accuracy of all tasks is better with the Gaussian Sampling than without. Similar results are observed for the CUB dataset, as seen in Table.~\ref{cub-runs}. From the tables, it is clear that with the addition of a single log function, the accuracy can improve by $\sim$0.5\%. This is a significant gain while using 5x lesser sampled data. All the results shown have been obtained after sampling 750 datapoints in total for the Gaussian sampling method, and 750 x 5 = 3750 datapoints for the distribution calibration method.

I also examine the effect of Gaussian Sampling with the variation of the number of datapoints sampled for each point in the support set. Fig.~\ref{mini-sample} shows the variation in average task accuracy for the miniImageNet dataset. Although the difference in accuracy is small, Gaussian sampling consistently outperforms distribution calibration \cite{free-lunch}.
Fig.~\ref{cub-sample} shows a similar trend in the variation in average accuracy for the CUB dataset with different amounts of sampled data.

\begin{table}[]
\begin{tabular}{|l|r|r|r|}
\hline
     Trial   & \multicolumn{1}{l|}{Without GS (\%)} & \multicolumn{1}{l|}{With GS (\%)} & \multicolumn{1}{l|}{Difference (\%)} \\ \hline
1 & 82.733                             & 83.053                          & 0.320                               \\ \hline
2 & 83.053                             & 84.067                          & 1.014                             \\ \hline
3 & 84.067                             & 84.177                          & 0.11                             \\ \hline
4 & 83.987                             & 84.24                            & 0.253                             \\ \hline
5 & 84.187                             & 84.973                          & 0.786                             \\ \hline
Avg & 83.605                            & 84.102                         & 0.497                            \\ \hline
\end{tabular}

\caption{Accuracy on the query set of the MiniImageNet dataset. Each trial consists of 100 tasks, run with and without Gaussian Sampling (GS).}
\label{miniImageNet-runs}
\end{table}

\begin{table}[]
\begin{tabular}{|l|r|r|r|}
\hline
    Trial    & \multicolumn{1}{l|}{Without GS (\%)} & \multicolumn{1}{l|}{With GS (\%)} & \multicolumn{1}{l|}{Difference (\%)} \\ \hline
1 & 90.747                               & 91.093                            & 0.346                                \\ \hline
2 & 91.747                               & 92.4                              & 0.653                                \\ \hline
3 & 91.747                               & 92.213                            & 0.466                                \\ \hline
4 & 90.307                               & 90.88                             & 0.573                                \\ \hline
5 & 90.307                               & 90.813                            & 0.506                                \\ \hline
Avg & 90.9707                              & 91.48                             & 0.509                                \\ \hline
\end{tabular}
\caption{Accuracy on the query set of the CUB dataset. Each trial consists of 100 tasks, run with and without Gaussian Sampling (GS).}
\label{cub-runs}
\end{table}

\begin{figure}
    \centering
    \includegraphics[width=6cm]{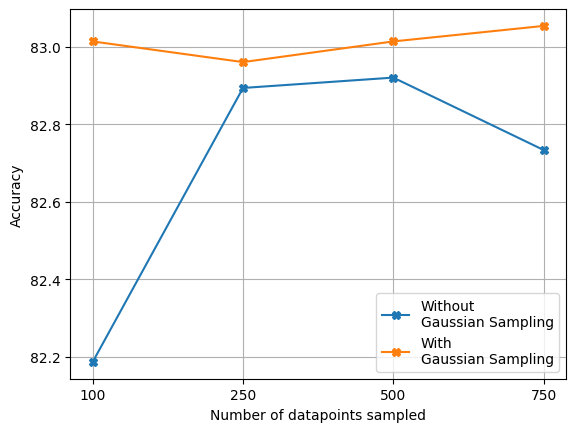}
    \caption{Variation in the accuracy with the number of sampled data per point in the support set, with/without Gaussian Sampling for the miniImageNet dataset. The number of datapoints sampled for the plot without Gaussian Sampling is 5x the value on the x-axis}
    \label{mini-sample}
\end{figure}

\begin{figure}
    \centering
    \includegraphics[width=6cm]{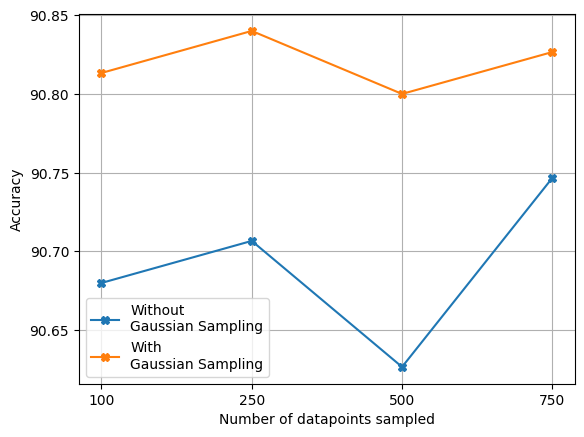}
    \caption{Variation in the accuracy with the number of sampled data per point in the support set, with/without Gaussian Sampling for the CUB dataset. The number of datapoints sampled for the plot without Gaussian Sampling is 5x the value on the x-axis}
    \label{cub-sample}
\end{figure}

\section{Conclusion}
\label{conclusion}

In this paper, I propose a new method to induce ``normality" within experimental data, called the Log-Tukey transform. By transforming data sampled from various distributions, I show the effectiveness of this transform in making data more Gaussian-like as compared to the existing methods. Further, I employ this transform to sample Gaussian representations in a few-shot learning method, and show significant incremental gains while reducing the amount of computation. I also demonstrate the generality and utility of the method by conducting experiments on datasets of varied granularity, and with different amount of sampled data. A gain of $\sim$0.5\% in accuracy by the addition of a single logarithm function seems like a good bargain. A possible direction for future work would be to examine the effectiveness of the Log-Tukey transform in other scenarios where Gaussian priors and sampling multi-variate Gaussians from experimental data are involved! 


{\small
\bibliographystyle{ieee_fullname}
\bibliography{egpaper_final}

\begin{thebibliography}{10}\itemsep=-1pt

\bibitem{tukey}
{\em Exploratory Data Analysis}, pages 192--194.
\newblock Springer New York, New York, NY, 2008.

\bibitem{box-cox}
G.~E.~P. Box and D.~R. Cox.
\newblock An analysis of transformations.
\newblock {\em Journal of the Royal Statistical Society: Series B
  (Methodological)}, 26(2):211--243, 1964.

\bibitem{metafree}
Kuilin Chen and Chi-Guhn Lee.
\newblock Meta-free few-shot learning via representation learning with weight
  averaging, 2022.

\bibitem{cub-split}
Wei-Yu Chen, Yen-Cheng Liu, Zsolt Kira, Yu-Chiang~Frank Wang, and Jia-Bin
  Huang.
\newblock A closer look at few-shot classification, 2020.

\bibitem{maml}
Chelsea Finn, Pieter Abbeel, and Sergey Levine.
\newblock Model-agnostic meta-learning for fast adaptation of deep networks,
  2017.

\bibitem{iris}
R.~A. Fisher.
\newblock {Iris}.
\newblock UCI Machine Learning Repository, 1988.
\newblock {DOI}: https://doi.org/10.24432/C56C76.

\bibitem{log-transform}
Loet Leydesdorff and Stephen Bensman.
\newblock Classification and powerlaws: The logarithmic transformation, 2009.

\bibitem{Gaussian-prototypes}
Jinfu Lin, Junmin Shen, and Xiaojian He.
\newblock Gaussian prototype rectification for few-shot image recognition.
\newblock In {\em 2021 International Joint Conference on Neural Networks
  (IJCNN)}, pages 1--8, 2021.

\bibitem{global-representations}
Tiange Luo, Aoxue Li, Tao Xiang, Weiran Huang, and Liwei Wang.
\newblock Few-shot learning with global class representations, 2019.

\bibitem{feature-extractor}
Puneet Mangla, Mayank Singh, Abhishek Sinha, Nupur Kumari, Vineeth~N
  Balasubramanian, and Balaji Krishnamurthy.
\newblock Charting the right manifold: Manifold mixup for few-shot learning,
  2020.

\bibitem{miniImagenet}
Sachin Ravi and Hugo Larochelle.
\newblock Optimization as a model for few-shot learning.
\newblock In {\em International Conference on Learning Representations}, 2017.

\bibitem{imagenet}
Olga Russakovsky, Jia Deng, Hao Su, Jonathan Krause, Sanjeev Satheesh, Sean Ma,
  Zhiheng Huang, Andrej Karpathy, Aditya Khosla, Michael Bernstein,
  Alexander~C. Berg, and Li Fei-Fei.
\newblock Imagenet large scale visual recognition challenge, 2015.

\bibitem{protonet}
Jake Snell, Kevin Swersky, and Richard~S. Zemel.
\newblock Prototypical networks for few-shot learning, 2017.

\bibitem{good-representation}
Yonglong Tian, Yue Wang, Dilip Krishnan, Joshua~B. Tenenbaum, and Phillip
  Isola.
\newblock Rethinking few-shot image classification: a good embedding is all you
  need?, 2020.

\bibitem{compositional-representations}
Pavel Tokmakov, Yu-Xiong Wang, and Martial Hebert.
\newblock Learning compositional representations for few-shot recognition,
  2019.

\bibitem{wasserstein}
C{\'e}dric Villani.
\newblock Optimal transport: Old and new.
\newblock 2008.

\bibitem{nk-shot}
Oriol Vinyals, Charles Blundell, Timothy Lillicrap, koray kavukcuoglu, and Daan
  Wierstra.
\newblock Matching networks for one shot learning.
\newblock In D. Lee, M. Sugiyama, U. Luxburg, I. Guyon, and R. Garnett,
  editors, {\em Advances in Neural Information Processing Systems}, volume~29.
  Curran Associates, Inc., 2016.

\bibitem{cub}
C. Wah, S. Branson, P. Welinder, P. Perona, and S. Belongie.
\newblock The caltech-ucsd birds-200-2011 dataset.
\newblock Technical Report CNS-TR-2011-001, California Institute of Technology,
  2011.

\bibitem{free-lunch}
Shuo Yang, Lu Liu, and Min Xu.
\newblock Free lunch for few-shot learning: Distribution calibration, 2021.

\bibitem{yeo-johnson}
In‐Kwon Yeo and Richard~A. Johnson.
\newblock {A new family of power transformations to improve normality or
  symmetry}.
\newblock {\em Biometrika}, 87(4):954--959, 12 2000.

\end{thebibliography}
}

\end{document}